\documentclass[runningheads]{llncs}
\usepackage{graphicx}
\usepackage[table]{xcolor}

\usepackage{xcolor}

\usepackage{threeparttable/threeparttable}
\usepackage{epsfig}
\usepackage{amsmath}
\usepackage{amssymb}
\usepackage{array}
\usepackage{multirow,makecell,multicol}
\usepackage{float}

\usepackage[pagebackref=true,breaklinks=true,letterpaper=true,colorlinks,bookmarks=false]{hyperref}

\newcolumntype{P}[1]{>{\centering\arraybackslash}p{#1}}

\newcommand{\ie}{\mbox{\emph{i.e.,\ }}}

\newenvironment{changemargin}[2]{%
\begin{list}{}{%
\setlength{\topsep}{0pt}%
\setlength{\leftmargin}{#1}%
\setlength{\rightmargin}{#2}%
\setlength{\listparindent}{\parindent}%
\setlength{\itemindent}{\parindent}%
\setlength{\parsep}{\parskip}%
}%
\item[]}{\end{list}}

\def\figurename{Fig.}
\def\sectionname{Sec.}
\def\tablename{Table}

\newenvironment{mhblue}{\color{black}}{}
\definecolor{iblue}{rgb}{0.06, 0.75, 1.0}
\begin{document}

\title{Learning Semantics-enriched Representation via\\Self-discovery, Self-classification, and Self-restoration}
\titlerunning{Semantic Genesis}
%
\author{Fatemeh Haghighi\inst{1} \and
Mohammad Reza Hosseinzadeh Taher\inst{1} \and
Zongwei Zhou\inst{1} \and
Michael B. Gotway\inst{2} \and
Jianming Liang\inst{1}}
%


\authorrunning{F. Haghighi et al.}
%

\institute{Arizona State University, Tempe AZ 85281, USA 
\email{\{fhaghigh,mhossei2,zongweiz,jianming.liang\}@asu.edu} \and
 Mayo Clinic, Scottsdale AZ 85259, USA\\
\email{Gotway.Michael@mayo.edu}}
\maketitle              

\vspace{-250pt}

\begin{changemargin}{-5cm}{-4cm} 
\begin{center}
\noindent\textbf{ \textcolor{red}{ This is the full version of our MICCAI-2020 paper on Semantic Genesis with the Supplementary Materials}\\[15pt]}
\end{center} 
\end{changemargin}

\begin{changemargin}{-2.7cm}{-2.7cm} 
   \noindent \textcolor{blue}{\small \textbf{Please cite the paper as F. Haghighi, M. R. Hosseinzadeh Taher, Z. Zhou, M. B. Gotway, and J. Liang. Learning Semantics-enriched Representation via Self-discovery, Self-classification, and Self-restoration. International Conference on Medical Image Computing and Computer-Assisted Intervention (MICCAI), 2020. }}
\end{changemargin}

\vspace{170pt}

\begin{abstract}
Medical images are naturally associated with rich semantics about the human anatomy, reflected in an abundance of recurring anatomical patterns, offering unique potential to foster deep semantic representation learning and yield semantically more powerful models for different medical applications. But how exactly such strong yet free semantics embedded in medical images can be harnessed for self-supervised learning remains largely unexplored. To this end, we train deep models to learn semantically enriched visual representation by self-discovery, self-classification, and self-restoration of the anatomy underneath medical images, resulting in a semantics-enriched, general-purpose, pre-trained 3D model, named Semantic Genesis.
We examine our Semantic Genesis with all the publicly-available pre-trained models, by either self-supervision or fully supervision, on the six distinct target tasks, covering both classification and segmentation in various medical modalities (\ie CT, MRI, and X-ray). Our extensive experiments demonstrate that Semantic Genesis significantly exceeds all of its 3D counterparts as well as the {\em de facto} ImageNet-based transfer learning in 2D.
This performance is attributed to our novel self-supervised learning framework, encouraging deep models to learn compelling semantic representation from abundant anatomical patterns resulting from consistent anatomies embedded in medical images. Code and pre-trained Semantic Genesis {\mhblue are} available at 
\href{https://github.com/JLiangLab/SemanticGenesis}{https://github.com/JLiangLab/SemanticGenesis}.

\end{abstract}

\section{{\mhblue Introduction}}
\noindent Self-supervised learning methods aim to learn general image representation from unlabeled data; naturally, a crucial question in self-supervised learning is how to ``extract'' proper supervision signals from the unlabeled data directly.
In large part, self-supervised learning approaches involve predicting some hidden properties of the data, such as  {\mhblue colorization~\cite{larsson2016learning,larsson2017colorization},  jigsaw~\cite{kim2018learning,noroozi2016unsupervised}, and rotation~\cite{feng2019self,gidaris2018unsupervised}.} However, most of the prominent methods were derived in the context of natural images, without considering the unique properties {\mhblue of medical images}.

\begin{figure}[t]
\centering
 \includegraphics[width=0.97\linewidth]{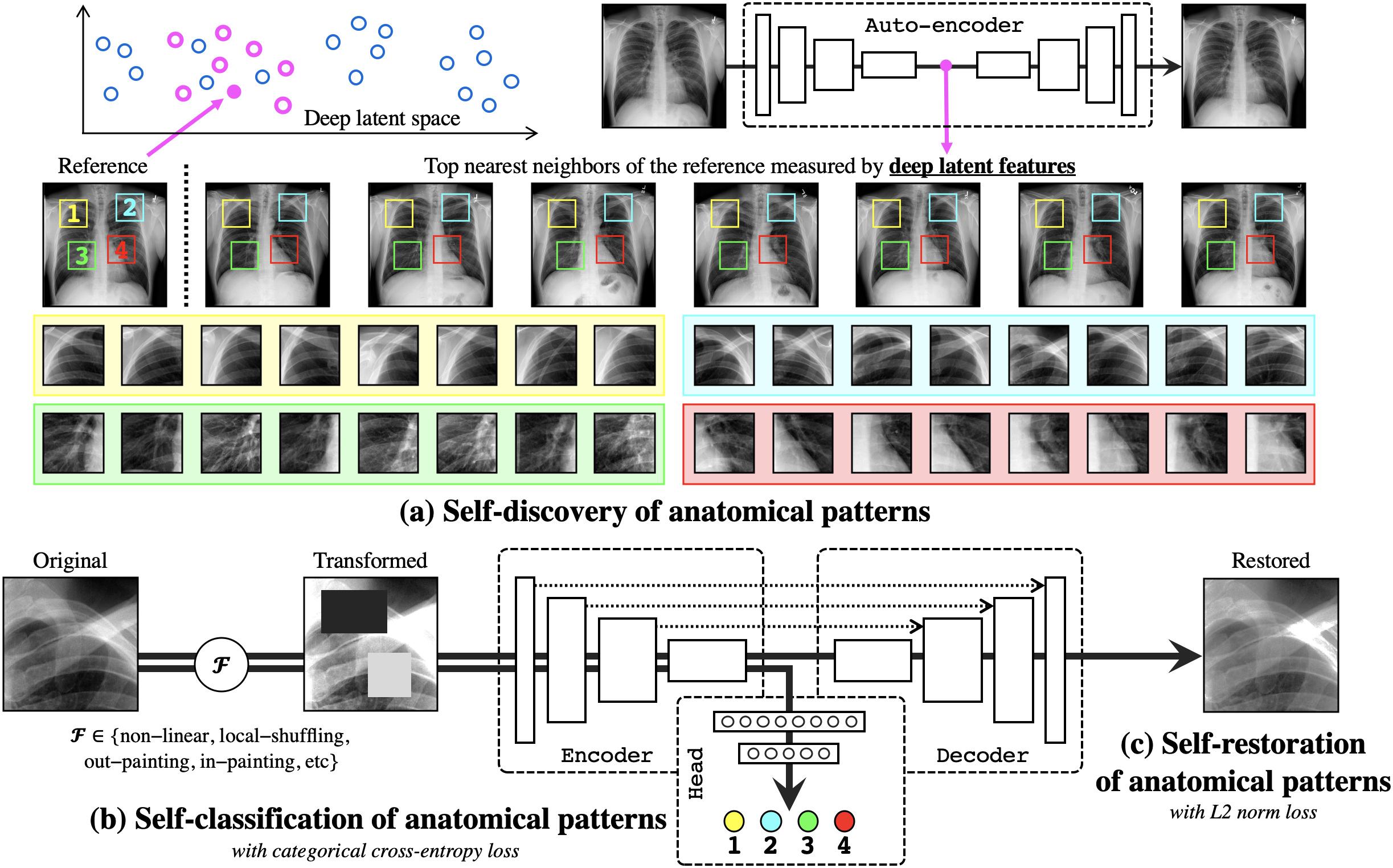}
\caption{
Our self-supervised learning framework consists of (a) self-discovery, (b) self-classification, and (c) self-restoration of anatomical patterns, resulting in semantics-enriched pre-trained models---Semantic Genesis---an encoder-decoder structure with skip connections in between and a classification head at the end of the encoder.
Given a random reference patient, we find similar patients based on deep latent features, crop anatomical patterns from random yet fixed coordinates, and assign pseudo labels to the crops according to their coordinates. For simplicity and clarity, we illustrate our idea with four coordinates in X-ray images as an example.
The input to the model is a transformed anatomical pattern crop, and the model is trained to classify the pseudo label and to recover the original crop\iffalse (detailed in~\sectionname~\ref{sec:method})\fi. \iffalse By doing so\fi {\mhblue Thereby}, the model aims to acquire semantics-enriched representation, producing more powerful application-specific target models\iffalse (detailed in~\sectionname~\ref{sec:results})\fi.}
\label{fig:method}
\end{figure}

In medical imaging, it is required to follow protocols for defined clinical purposes, therefore generating images of similar anatomies across patients and yielding recurrent anatomical patterns across images (see~\figurename~\ref{fig:method}a). 
These recurring patterns are associated with rich semantic knowledge about the human body, thereby offering great potential to foster deep semantic representation learning and produce more powerful models for various medical applications. However, it remains an unanswered question: \textit{How to exploit the deep semantics associated with recurrent anatomical patterns embedded in medical images to enrich representation learning?}

To answer this question, we present a novel self-supervised learning framework, which enables the capture of semantics-enriched representation from unlabeled medical image data, resulting in a set of powerful pre-trained models. 
{\mhblue We  call  our  pre-trained
models \textbf{Semantic Genesis}, because they represent a significant advancement from Models Genesis~\cite{zhou2019models} by introducing two {\em novel} components: self-discovery and self-classification of  the  anatomy  underneath  medical  images (detailed in~\sectionname~\ref{sec:method}). Specifically, our {\em unique} self-classification branch, with a small computational overhead, compels the model to learn semantics from consistent and recurring anatomical patterns discovered during the self-discovery phase, while Models  Genesis learns representation from random sub-volumes with no semantics as no semantics can be discovered from random sub-volumes. By explicitly employing the strong yet free semantic supervision signals, Semantic Genesis distinguishes itself from all other existing works, including colorization of colonoscopy images~\cite{ross2018exploiting},  context restoration~\cite{chen2019self}, Rubik's cube recovery~\cite{Zhuang2019Self}, and predicting anatomical positions within MR images~\cite{Bai2019Self}.

As evident in~\sectionname~\ref{sec:results}, our extensive experiments demonstrate that (1) learning semantics through our two innovations significantly enriches existing self-supervised learning approaches~\cite{chen2019self,pathak2016context,zhou2019models}, boosting target tasks performance dramatically (see~\figurename~\ref{fig:result_semantics_included}); (2) Semantic Genesis provides more generic and transferable feature representations in comparison to not only its self-supervised learning counterparts, but also (fully) supervised pre-trained 3D models (see~\tablename~\ref{tab:3d_semantic_genesis_top}); and Semantic Genesis significantly surpasses any 2D approaches (see~\figurename~\ref{fig:2d_semantic_genesis_top}).
}

This performance is ascribed to the semantics derived from the consistent and recurrent anatomical patterns, that not only can be automatically discovered from medical images but can also serve as strong yet free supervision signals for deep models to learn more semantically enriched representation automatically via self-supervision.

\section{Semantic Genesis}
\label{sec:method}

\figurename~\ref{fig:method} presents our self-supervised learning framework, which enables training Semantic Genesis from scratch on unlabeled medical images. 
Semantic Genesis is conceptually simple: an encoder-decoder structure with skip connections in between and a classification head at the end of the encoder. 
The objective for the model is to learn different sets of semantics-enriched representation from multiple perspectives. 
In doing so, our proposed framework consists of three important components: 1) self-discovery of anatomical patterns from similar patients; 2) self-classification of the patterns; and 3) self-restoration of the transformed patterns.
Specifically, once the self-discovered anatomical pattern set is built, we jointly train the classification and restoration branches together in the model.

\medskip
\noindent{\bf {\mhblue 1) Self-discovery of anatomical patterns:}}
\label{sec:method_discovery}
We begin by building a set of anatomical patterns from medical images, as illustrated in~\figurename~\ref{fig:method}a. To extract deep features of each (whole) patient scan, we first train an auto-encoder network with training data, which learns an identical mapping from scan to itself. Once trained, the latent representation vector from the auto-encoder can be used as an indicator of each patient. We randomly anchor one patient as a reference and search for its nearest neighbors through the entire dataset by computing the $L2$ distance of the latent representation vectors, resulting in a set of semantically similar patients. As shown in~\figurename~\ref{fig:method}a, due to the consistent and recurring anatomies across these patients, that is, each coordinate contains a unique anatomical pattern, it is feasible to extract similar anatomical patterns according to the coordinates. Hence, we crop patches/cubes (for 2D/3D images) from $C$ number of random but fixed coordinates across this small set of discovered patients, which share similar semantics. Here we compute similarity in patient-level rather than pattern-level to ensure the balance between the diversity and consistency of anatomical patterns. Finally, we assign pseudo labels to these patches/cubes based on their coordinates, resulting in a new dataset, wherein each {\mhblue patch/cube} is associated with one of the $C$ classes. Since the coordinates are randomly selected in the reference patient, { \mhblue some of the anatomical patterns \iffalse in most of the classes\fi may not be very meaningful} for radiologists, yet these patterns are still associated with rich local semantics of the human body. For example, in~\figurename~\ref{fig:method}a, four pseudo labels are defined randomly in the reference patient (top-left most), but as seen, they carry local information of (1) anterior ribs 2--4, (2) anterior ribs 1--3, (3) right pulmonary artery, and (4) LV. Most importantly, by repeating the above self-discovery process, enormous anatomical patterns associated with their pseudo labels can be automatically generated for representation learning in the following stages (refer to Appendix~\sectionname~\ref{appendix_self_discovery}).

\medskip
 \noindent{\bf {\mhblue 2) Self-classification of anatomical patterns:} } 
\label{sec:method_classification}
After self-discovery of  a set of anatomical patterns, we formulate the representation learning as a $C$-way multi-class classification task.
The goal is to encourage models to learn from the recurrent anatomical patterns across patient images, fostering a deep semantically enriched representation.
As illustrated in~\figurename~\ref{fig:method}b, the classification branch encodes the input anatomical pattern into a latent space, followed by a sequence of fully-connected (\textit{fc}) layers, and predicts the pseudo label associated with the pattern.
To classify the anatomical patterns, we adopt categorical cross-entropy loss function: $\mathcal{L}_{cls}=-\frac{1}{N}\sum_{b=1}^{N}\sum_{c=1}^{C}\mathcal{Y}_{bc}\log  {\mathcal{P}_{bc}}$, where $N$ denotes the batch size; $C$ denotes the number of classes; $\mathcal{Y}$ and $\mathcal{P}$ represent the ground truth (one-hot pseudo label vector) and the prediction, respectively.

\medskip
\noindent{\bf {\mhblue 3) Self-restoration of anatomical patterns:}}
\label{sec:method_restoration}
The objective of self-restoration is for the model to learn different sets of visual representation by recovering original anatomical patterns from the transformed ones.
{\mhblue We adopt the \iffalse same  four\fi transformations \iffalse as  those initially\fi proposed in Models Genesis~\cite{zhou2019models},} \ie non-linear, local-shuffling, out-painting, and in-painting {\mhblue (refer to Appendix~\sectionname~\ref{sec:distortion})}. 
As shown in~\figurename~\ref{fig:method}c, the restoration branch encodes the input transformed anatomical pattern into a latent space and decodes back to the original resolution, with an aim to recover the original anatomical pattern from the transformed one.
To let Semantic Genesis restore the transformed anatomical patterns, we compute $L2$ distance between original pattern and reconstructed pattern as loss function: $\mathcal{L}_{rec}=\frac{1}{N}\sum_{i=1}^{N}\|\mathcal{X}_i - \mathcal{X}_i'\|_2$, 
{\mhblue where $N$, $\mathcal{X}$ and $\mathcal{X}'$ denote the batch size, ground truth (original anatomical pattern) and reconstructed prediction, respectively.}

Formally, during training, we define a multi-task loss function on each transformed anatomical pattern as $\mathcal{L}=\lambda_{cls}\mathcal{L}_{cls}+\lambda_{rec}\mathcal{L}_{rec}$, where $\lambda_{cls}$ and $\lambda_{rec}$ regulate the weights of classification and reconstruction losses, respectively. 
Our definition of $\mathcal{L}_{cls}$ allows the model to learn more semantically enriched representation.
The definition of $\mathcal{L}_{rec}$ encourages the model to learn from multiple perspectives by restoring original images from varying image deformations.
Once trained, the encoder alone can be fine-tuned for target \textit{classification} tasks; while the encoder and decoder together can be fine-tuned for target \textit{segmentation} tasks  to fully utilize the advantages of the pre-trained models on the target tasks.

\begin{table}[t]
\begin{center}

\begin{threeparttable}
\caption{
We evaluate the learned representation by fine-tuning it for six  publicly-available medical imaging applications including 3D and 2D image classification and segmentation tasks, across diseases, organs, datasets, and modalities.
}
\label{tab:datasets}
{\mhblue 
\begin{tabular}{p{0.07\linewidth}p{0.175\linewidth} P{0.11\linewidth}p{0.192\linewidth}p{0.39\linewidth}}
\hline
Code$^{a}$ &  Object & Modality & Dataset & Application \\
\hline
\texttt{NCC}  &Lung Nodule &CT & {\scriptsize LUNA-2016~\cite{setio2017validation} }& {\scriptsize Nodule false positive reduction}  \\
\texttt{NCS}  &Lung Nodule  &CT & {\scriptsize LIDC-IDRI~\cite{armato2011lung} }& {\scriptsize Lung nodule segmentation } \\
\texttt{LCS}  &Liver &CT & {\scriptsize LiTS-2017~\cite{bilic2019liver} } & {\scriptsize Liver segmentation} \\
\texttt{BMS}  &Brain Tumor &MRI & {\scriptsize BraTS2018~\cite{bakas2018identifying} } & {\scriptsize Brain Tumor Segmentation} \\
\texttt{DXC}  & Chest Diseases&X-ray & {\scriptsize ChestX-Ray14~\cite{wang2017chestx} } & {\scriptsize Fourteen chest diseases classification} \\
\texttt{\texttt{PXS}}&Pneumothorax &X-ray& {\scriptsize SIIM-ACR-2019~\cite{PNEchallenge}} & {\scriptsize Pneumothorax Segmentation} \\
\hline
\end{tabular}
}
\begin{tablenotes}
        \item $^{a}$ The first letter denotes the object of interest (``\texttt{N}'' for lung nodule, ``\texttt{L}'' for liver, etc); the second letter denotes the modality (``\texttt{C}'' for CT, ``\texttt{X}'' for X-ray, ``\texttt{M}'' for MRI);  the last letter denotes the task (``\texttt{C}'' for classification, ``\texttt{S}'' for segmentation).
\end{tablenotes}
\end{threeparttable}
\end{center}
\end{table}

\section{Experiments}
\label{sec:experiments}
\noindent\textbf{Pre-training Semantic Genesis:}
Our Semantic Genesis 3D and 2D are self-supervised pre-trained from 623 CT scans in LUNA-2016~\cite{setio2017validation} (same as the publicly released Models Genesis) and 75,708 X-ray images from ChestX-ray14~\cite{wang2017comparison} datasets, respectively. Although Semantic Genesis is trained from only unlabeled images, we do not use all the images in those datasets to avoid test-image leaks between proxy and target tasks. In the self-discovery process, we select top $K$ most similar cases with the reference patient, according to the deep features computed from the pre-trained auto-encoder. To strike a balance between diversity and consistency of the anatomical patterns, we empirically set $K$ to 200/1000 for 3D/2D pre-training based on the dataset size. We set $C$ to 44/100 for 3D/2D images so that the anatomical patterns can largely cover the entire image while avoiding too much overlap with each other. For each random coordinate, we extract multi-resolution cubes/patches,  then resize them all to 64$\times$64$\times$32 and 224$\times$224 for 3D and 2D, respectively; finally, we assign $C$ pseudo labels to the cubes/patches based on their coordinates. For more details in implementation and meta-parameters, please refer to our publicly released code.

\smallskip
\noindent\textbf{Baselines and implementation:}
{\mhblue \tablename~\ref{tab:datasets} summarizes the target tasks and datasets.} Since most self-supervised learning methods are initially proposed in 2D, we have extended two most representative ones~\cite{chen2019self,pathak2016context} into their 3D version for a fair comparison. 
{\mhblue Also, we compare Semantic Genesis with Rubik's cube~\cite{Zhuang2019Self}, the most recent multi-task self-supervised learning method for 3D medical imaging.}
In addition, we have examined publicly available pre-trained models for 3D transfer learning in medical imaging, including NiftyNet~\cite{gibson2018niftynet}, MedicalNet~\cite{chen2019med3d}, Models Genesis~\cite{zhou2019models}, and Inflated 3D (I3D)~\cite{carreira2017quo} that has been successfully transferred to 3D lung nodule detection~\cite{ardila2019end}, as well as ImageNet models, the most influential weights initialization in 2D target tasks. 
{\mhblue 3D U-Net\footnote{\label{foot:3dunet}3D U-Net: \href{https://github.com/ellisdg/3DUnetCNN}{github.com/ellisdg/3DUnetCNN}}/U-Net\footnote{\label{foot:densenet121}Segmentation Models: \href{https://github.com/qubvel/segmentation_models}{github.com/qubvel/segmentation\_models}} architectures used in 3D/2D applications,} have been modified by appending fully-connected layers to end of the encoders.
In proxy tasks, we set $\lambda_{rec}=1$ and $\lambda_{cls}=0.01$. Adam with a learning rate of 0.001 is used for optimization. 
We first train classification branch for 20 epochs, then jointly train the entire model for both classification and restoration tasks. 
For CT target tasks, we investigate the capability of both 3D volume-based solutions and 2D slice-based solutions, where the 2D representation is obtained by extracting axial slices from volumetric datasets. 
For all applications, we run each method 10 times on the target task and report the average, standard deviation, and further present statistical analyses based on independent two-sample \textit{t}-test.

\begin{figure}[t]
\begin{center}
\includegraphics[width=0.95\linewidth]{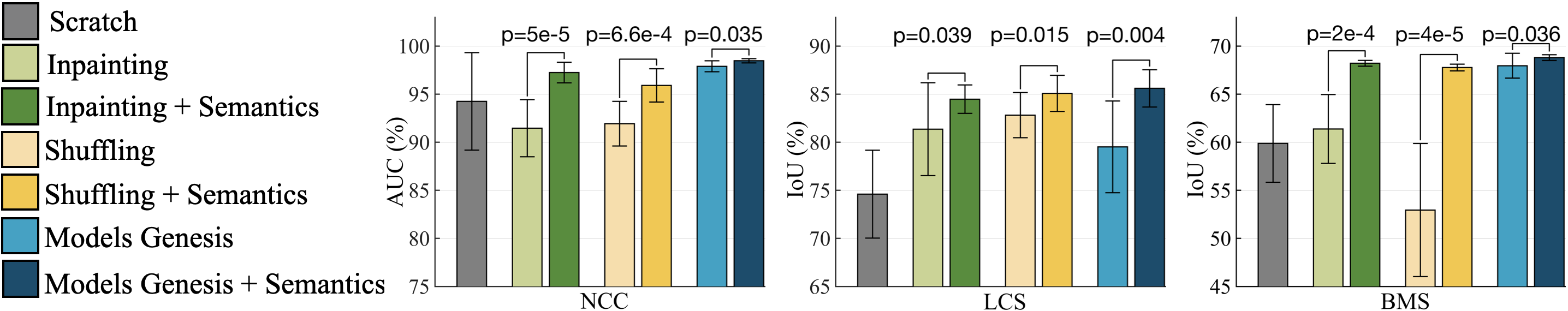}
\end{center}
\caption{
   With and without semantics-enriched representation in the self-supervised learning approaches contrast a substantial ($p<0.05$) performance difference on target classification and segmentation tasks.
    By introducing self-discovery and self-classification, we enhance semantics in three most recent self-supervised learning advances (\ie image in-painting~\cite{pathak2016context}, patch-shuffling~\cite{chen2019self}, and Models Genesis~\cite{zhou2019models}).
}
\label{fig:result_semantics_included}
\end{figure}

\section{Results}
\label{sec:results}
\noindent\textbf{Learning semantics enriches existing self-supervised learning approaches:}
{\mhblue Our proposed self-supervised learning scheme should be considered as an {\em add-on}, which can be added to and boost existing self-supervised learning methods. \iffalse As shown in~\figurename~\ref{fig:result_semantics_included},\fi Our results in~\figurename~\ref{fig:result_semantics_included} }indicate that by simply incorporating the anatomical patterns with representation learning, the semantics-enriched models consistently outperform each and every existing self-supervised learning method~\cite{pathak2016context,chen2019self,zhou2019models}.
Specifically, the semantics-enriched representation learning achieves performance gains by 5\%, 3\%, and 1\% in \texttt{NCC}, compared with the original in-painting, patch-shuffling, and Models Genesis, respectively; and the performance improved by 3\%, 2\%, and 6\% in \texttt{LCS} and  6\%, 14\%, and 1\% in \texttt{BMS}. We conclude that our proposed self-supervised learning scheme, by autonomously discovering and classifying anatomical patterns, learns a unique and complementary visual representation in comparison with that of an image restoration task. Thereby, due to this combination, the models are enforced to learn from multiple perspectives, especially from the consistent and recurring anatomical structure, resulting in more powerful image representation. 

\begin{table*}[t]
\begin{center}
\begin{threeparttable}
\scriptsize
\caption{
Semantic Genesis outperforms learning 3D models from scratch, three competing publicly available (fully) supervised pre-trained 3D models, and four self-supervised learning approaches in four target tasks. For every target task, we report the mean and standard deviation (mean$\pm$s.d.) across ten trials and further perform independent two sample $t$-test between the best (bolded) vs. others and  highlighted boxes in blue when they are not statistically significantly different at $p=0.05$ level.
}
\label{tab:3d_semantic_genesis_top}

\begin{tabular*}{\textwidth}
{p{0.15\linewidth} p{0.21\linewidth} P{0.15\linewidth} 
P{0.15\linewidth}P{0.15\linewidth}P{0.15\linewidth}}
\hline
Pre-training& Initialization&
\texttt{NCC} (AUC\%) & \texttt{LCS} (IoU\%) & \texttt{NCS} (IoU\%) &\texttt{BMS}$^\ddagger$ (IoU\%)\\
\hline
& Random&94.25$\pm$5.07& 74.60$\pm$4.57 & 74.05$\pm$1.97 & 59.87$\pm$4.04\\
\hline
\multirow{3}{*}{ Supervised} & NiftyNet~\cite{gibson2018niftynet} &94.14$\pm$4.57&83.23$\pm$1.05  &52.98$\pm$2.05& 60.78$\pm$1.60\\
&MedicalNet~\cite{chen2019med3d}& 95.80$\pm$0.51& 83.32$\pm$0.85 & 75.68$\pm$0.32 & 66.09$\pm$1.35\\
&Inflated 3D (I3D)~\cite{carreira2017quo}&98.26$\pm$0.27 &70.65$\pm$4.26&71.31$\pm$0.37 & 67.83$\pm$0.75 \\
\hline
\multirow{7}{*}{Self-supervised} & Autoencoder& 88.43$\pm$10.25 & 78.16$\pm$2.07 & 75.10$\pm$0.91& 56.36$\pm$5.32\\
&In-painting~\cite{pathak2016context}& 91.46$\pm$2.97 & 81.36$\pm$4.83 & 75.86$\pm$0.26 & 61.38$\pm$3.84 \\
&Patch-shuffling~\cite{chen2019self}&91.93$\pm$2.32 & 82.82$\pm$2.35 & 75.74$\pm$0.51&52.95$\pm$6.92\\
&Rubik's Cube~\cite{Zhuang2019Self}& 95.56$\pm$ 1.57& 76.07$\pm$ 0.20 & 70.37$\pm$1.13& 62.75$\pm$1.93 \\
\cline{2-6}
&Self-restoration~\cite{zhou2019models}& 98.07$\pm$0.59 & 78.78$\pm$3.11&  \cellcolor{iblue!30}\textbf{77.41$\pm$0.40} &67.96$\pm$1.29\\
&Self-classification& 97.41$\pm$0.32& 83.61$\pm$2.19 & 76.23$\pm$0.42& 66.02$\pm$0.83\\
&Semantic Genesis 3D& \cellcolor{iblue!30}\textbf{98.47$\pm$0.22}&  \cellcolor{iblue!30}\textbf{85.60$\pm$1.94 }& \cellcolor{iblue!30} 77.24$\pm$0.68 & \cellcolor{iblue!30}\textbf{68.80$\pm$0.30}\\
\hline
\end{tabular*}

\begin{tablenotes}
 \scriptsize
        \item $^\ddagger$  Models Genesis used only \textit{synthetic} images of BraTS-2013, however we examine \textit{real} and only MR Flair images for segmenting brain tumors, so the results are not submitted to BraTS-2018.
    \end{tablenotes}
    \end{threeparttable}
    \end{center}
\end{table*}

\medskip
\noindent\textbf{{\mhblue Semantic Genesis 3D provides more generic and transferable representations in comparison to publicly available pre-trained 3D models:} }
We have compared our Semantic Genesis 3D with the competitive publicly available pre-trained models, applied to four distinct 3D target medical applications.
Our statistical analysis in~\tablename~\ref{tab:3d_semantic_genesis_top} suggests three major results.
Firstly, compared to learning 3D models from scratch, fine-tuning from Semantic Genesis offers performance gains by at least 3\%, while also yielding more stable performances in all four applications.
Secondly, fine-tuning models from Semantic Genesis achieves significantly higher performances than those fine-tuned from other self-supervised approaches, in all four distinct 3D medical applications, \ie \texttt{NCC, \texttt{LCS}, \texttt{NCS}, and \texttt{BMS}}. {\mhblue In particular, Semantic Genesis surpasses Models Genesis, the state-of-the-art 3D pre-trained models created by image restoration based self-supervised learning, in three applications (\ie \texttt{NCC}, \texttt{LCS}, and \texttt{BMS}),} and offers equivalent performance in \texttt{NCS}.
Finally, even though our Semantic Genesis learns representation without using any human annotation, we still have examined it with 3D models pre-trained from full supervision, \ie MedicalNet, NiftyNet, and I3D. Without any bells and whistles, Semantic Genesis outperforms supervised pre-trained models in all four target tasks. Our results evidence that in contrast to other baselines, which show fluctuation in different applications, Semantic Genesis is consistently capable of generalizing well in all tasks even when the domain distance between source and target datasets is large (\ie \texttt{LCS} and \texttt{BMS} tasks).
Conversely, Semantic Genesis benefits explicitly from the deep semantic features enriched by self-discovering and self-classifying anatomical patterns embedded in medical images, and thus contrasts with any other existing 3D models pre-trained by either self-supervision or full supervision.

\begin{figure}[t]
\begin{center}
 \includegraphics[width=0.9\linewidth]{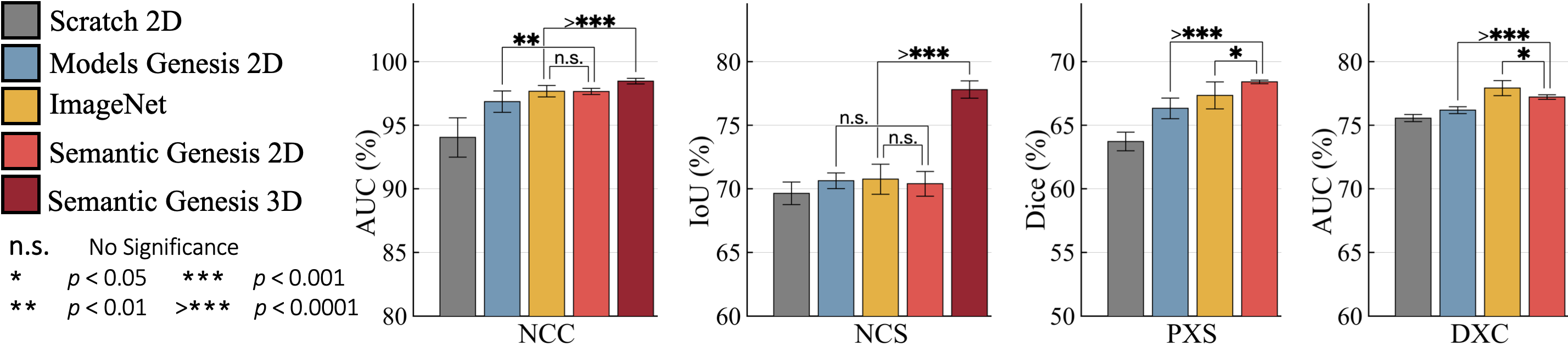}
\end{center}
   \caption{
   To solve target tasks in 3D medical modality (\texttt{NCC} and \texttt{NCS}), 3D approaches empowered by Semantic Genesis 3D, significantly outperforms any 2D slice-based approaches, including the state-of-the-art ImageNet models. For target tasks in 2D modality (\texttt{PXS} and \texttt{DXC}), Semantic Genesis 2D outperforms Models Genesis 2D and, noticeably, yields higher performance than ImageNet in \texttt{PXS}. 
   }
\label{fig:2d_semantic_genesis_top}
\end{figure}

\medskip
{\mhblue
\noindent\textbf{Semantic Genesis 3D  significantly surpasses any 2D approaches:}
To address the problem of limited annotation in volumetric medical imaging, one can reformulate and solve 3D imaging tasks in 2D~\cite{zhou2019models}. However, this approach may lose rich 3D anatomical  information  and  inevitably compromise  the  performance. Evidenced  by Fig. 3 (\texttt{NCC} and \texttt{NCS}), Semantic Genesis 3D outperforms all 2D solutions, including ImageNet models as well as downgraded Semantic Genesis 2D and Models Genesis  2D,  demonstrating  that  3D  problems  in  medical  imaging  demand  3D solutions.}
Moreover, as an ablation study, we examine our Semantic Genesis 2D with Models Genesis 2D (self-supervised) and ImageNet models (fully supervised) in four target tasks, covering classification and segmentation in CT and X-ray. Referring to \figurename~\ref{fig:2d_semantic_genesis_top}, Semantic Genesis 2D:  1) significantly surpasses training from scratch and Models Genesis 2D in all four and three applications, respectively; 2) outperforms ImageNet model in \texttt{PXS} and achieves the performance equivalent to ImageNet in \texttt{NCC} and \texttt{NCS}, which is a significant achievement because to date, all self-supervised approaches lag behind fully supervised training~\cite{hendrycks2019using,caron2019unsupervised,zhang2019aet}.

\medskip
\noindent\textbf{Self-classification  and  self-restoration  lead  to complementary representation:}
\label{sec:discussion}
\label{sec:discuss_loss}
In theory, our Semantic Genesis benefits from two sources: pattern classification and pattern restoration, so we further conduct an ablation study to investigate the effect of each isolated training scheme. 
Referring to~\tablename~\ref{tab:3d_semantic_genesis_top}, the combined training scheme (Semantic Genesis 3D) consistently offers  significantly higher and more stable performance compared to each of the isolated training schemes (self-restoration and self-classification)  in \texttt{NCS}, \texttt{LCS}, and \texttt{BMS}. Moreover, self-restoration and self-classification reveal better performances in four target applications, alternatingly.
We attribute their complementary  results  to the different visual representations that they have captured from each isolated pre-training scheme, leading to different behaviors in different target applications.
These complementary representations, in turn, confirm the importance of the unification of self-classification and self-restoration in our Semantic Genesis and its significance for medical imaging.

\section{Conclusion}
\label{sec:conclusion}
A key contribution of ours is designing a self-supervised learning framework that not only allows deep models to learn common visual representation from image data directly, but also leverages semantics-enriched representation from the consistent and recurrent anatomical patterns, one of a broad set of unique properties that medical imaging has to offer. Our extensive results  demonstrate that Semantic Genesis is superior to publicly available 3D models   pre-trained by either self-supervision or even full supervision, as well as ImageNet-based transfer learning in 2D. We attribute this outstanding results to the compelling deep semantics learned from abundant anatomical patterns resulted form consistent anatomies naturally embedded in medical images.

\subsubsection {Acknowledgments:} This research has been supported partially by ASU and Mayo Clinic through a Seed Grant and an Innovation Grant, and partially by the NIH under Award Number R01HL128785. The content is solely the responsibility of the authors and does not necessarily represent the official views of the NIH. We thank Zuwei Guo for implementing Rubik's cube, and Jiaxuan Pang for evaluating I3D. The content of this paper is covered by patents pending.

%
%
%
%

\newpage
\appendix
\section*{Appendix} 
\section{Visualizing the self-discovery process in Semantic Genesis}
\label{appendix_self_discovery}

\begin{figure*}[h]
\begin{center}
 \includegraphics[width=1\linewidth]{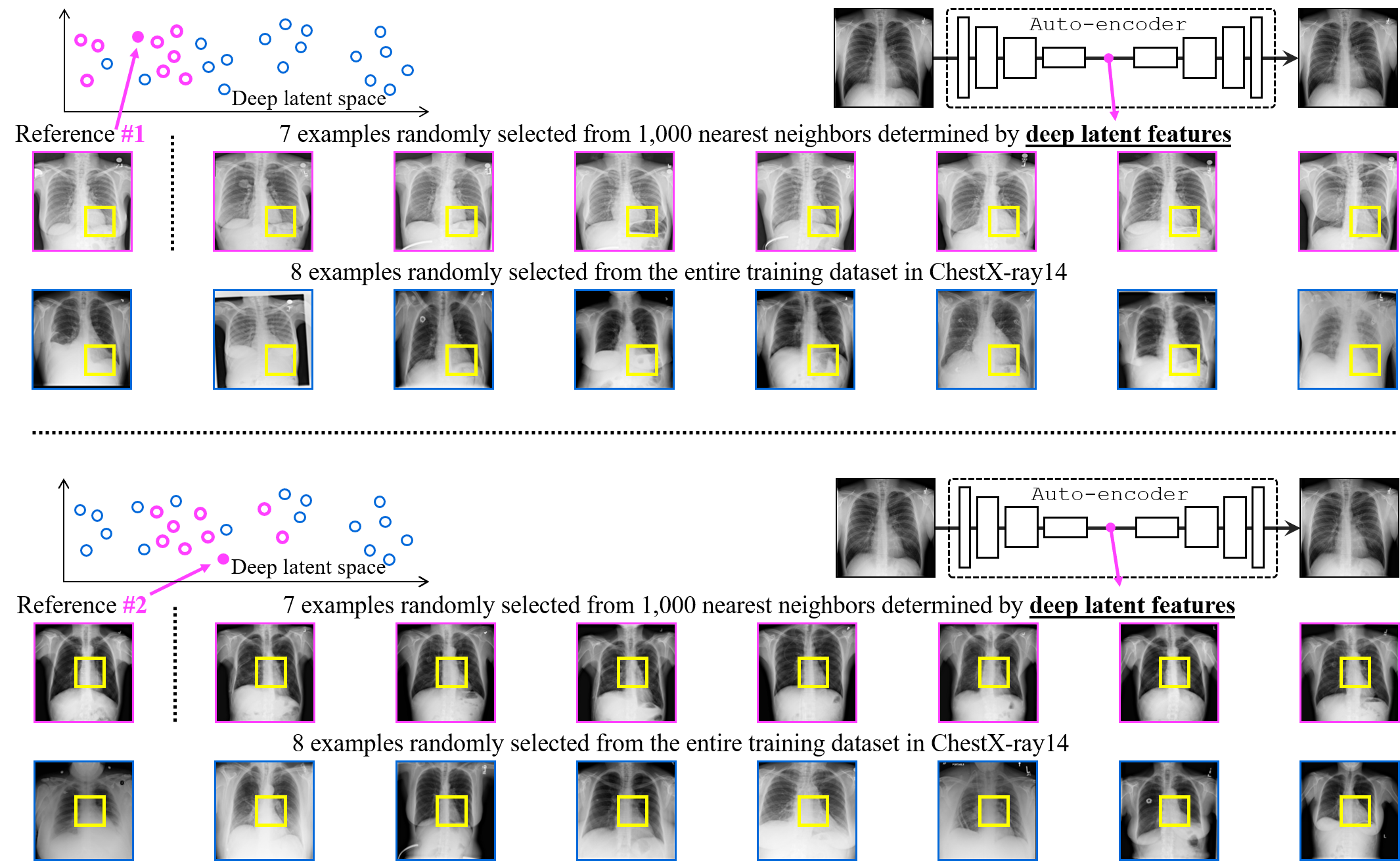}
\end{center}
\caption [LoF entry]{
Our self-discovery process aims to automatically discover similar anatomical patterns across patients, as illustrated in the yellow boxes within the patients framed in pink. Patches extracted at the same coordinate across patients may be very different (the yellow boxes within the patients framed in blue). We overcome this issue by first computing similarity at the patient level using the deep latent features from an auto-encoder and then selecting the top nearest neighbors (framed in pink) of the reference patient. Extracting anatomical patterns from these similar patients strikes a balance between consistency and diversity in pattern appearance for each anatomical pattern.
}
\label{fig:appendix_discover_patients}
\end{figure*}

\newpage
\section{Visualizing transformed anatomical patterns}
\label{sec:distortion}
\begin{figure*}[h]
\begin{center}
 \includegraphics[width=1\linewidth]{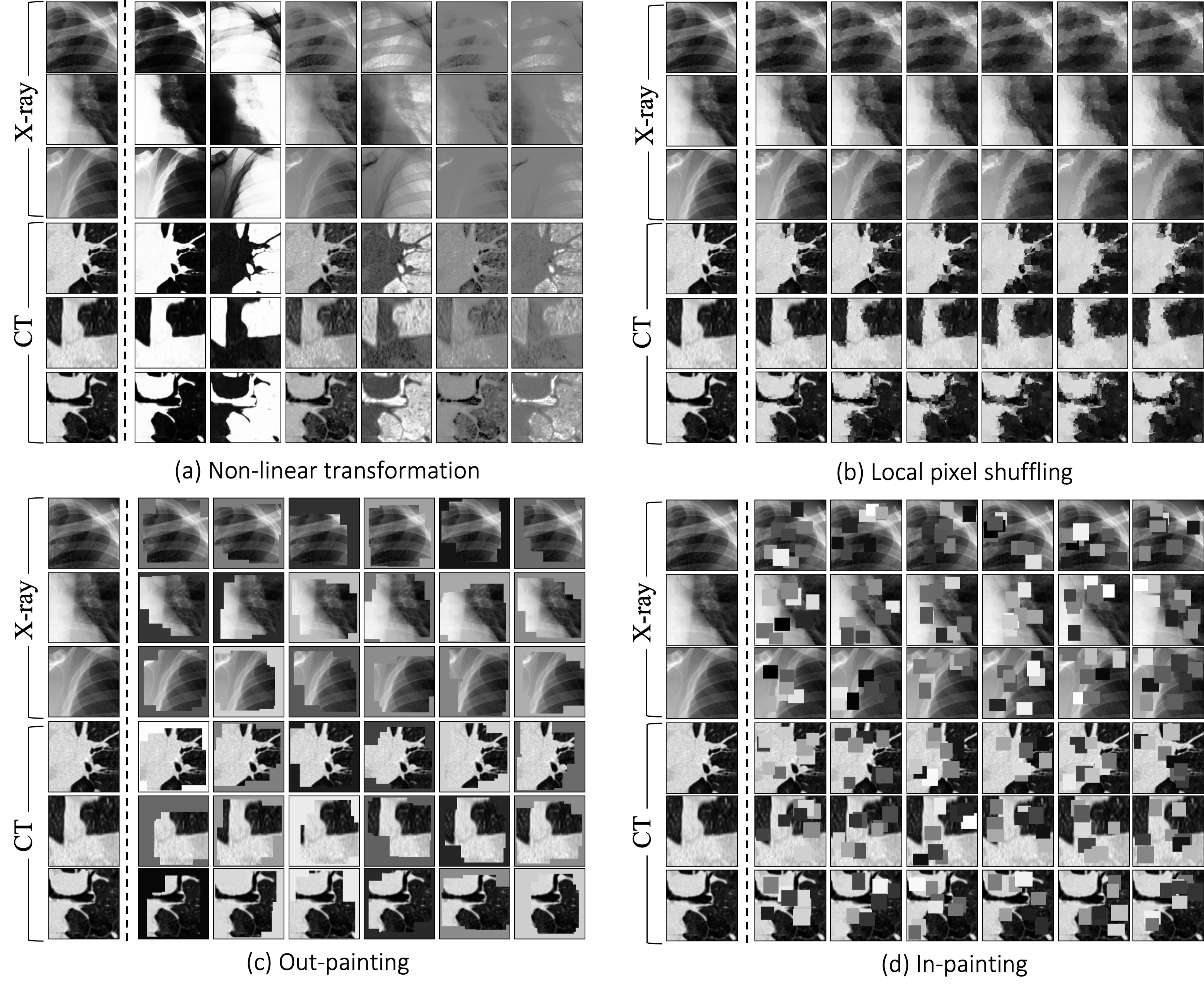}
\end{center}
\caption{
In the self-restoration process, Semantic Genesis aims to learn general-purpose visual representation by recovering original anatomical patterns from their transformed ones. We have adopted four image transformations as suggested in~\cite{zhou2019models}. To be self-contained, we provide three examples of anatomical patterns from CT slices and three from X-ray images. The original and transformed anatomical patterns are presented in Column 1 and Columns 2---7, respectively. Note that the original Models Genesis~\cite{zhou2019models} involve \textit{no} anatomical patterns but just random patches, while our Semantic Genesis benefits from the rich semantics associated with recurrent anatomical patterns embedded in medical images.
}
\label{fig:appendix_discover_patients}
\end{figure*}

\end{document}